\documentclass[letterpaper, 10 pt, conference]{ieeeconf}  

\IEEEoverridecommandlockouts                              

\usepackage{amsmath} 
\usepackage{amssymb}

\title{\LARGE \bf
Transformers for Object Detection in Large Point Clouds
}

\author{Felicia Ruppel$^{1,2}$, Florian Faion$^1$, Claudius Gl\"{a}ser$^1$ and Klaus Dietmayer$^2$
\thanks{$^{1}$Robert Bosch GmbH, Corporate Research, 71272 Renningen, Germany, 
        {\tt\small \{firstname.lastname\}@de.bosch.com}}%
\thanks{$^{2}$Institute of Measurement, Control and Microtechnology, Ulm University, Germany,
        {\tt\small \{firstname.lastname\}@uni-ulm.de}}%
}
\usepackage[noadjust]{cite}
\usepackage{graphicx}
\usepackage{subfig}
\usepackage{pgfplots}
\usepgfplotslibrary{fillbetween}
\pgfplotsset{
	max space between ticks=30pt,
	try min ticks=3,
	every axis/.style={
		axis y line=left,
		axis x line=bottom,
		axis line style={thick,->,>=latex, shorten >=-.4cm}
	},
	every axis plot/.append style={thick},
	tick style={black, thick}
}
\tikzset{
	semithick/.style={line width=1.1pt},
}
\usepackage{booktabs}
\usepackage{multirow}
\usepackage[colorinlistoftodos,color=pink]{todonotes} 
\usepackage{bm}
\presetkeys{todonotes}{inline}{}

\newcounter{todocounter}

\usepackage{fancyhdr}
\usepackage{ragged2e}

\fancypagestyle{FirstPage}{
	
	\lfoot{\noindent\fontsize{7.5}{8.5}\selectfont\justifying\textcopyright\ 2022 IEEE.  Personal use of this material is permitted.  Permission from IEEE must be obtained for all other uses, in any current or future media, including reprinting/republishing this material for advertising or promotional purposes, creating new collective works, for resale or redistribution to servers or lists, or reuse of any copyrighted component of this work in other works.}
	
}
\makeatletter
\let\NAT@parse\undefined
\makeatother
\usepackage{hyperref}

\begin{document}
\maketitle
\thispagestyle{empty}
\pagestyle{empty}

\begin{abstract}We present TransLPC, a novel detection model for large point clouds that is based on a transformer architecture. While object detection with transformers has been an active field of research, it has proved difficult to apply such models to point clouds that span a large area, e.g. those that are common in autonomous driving, with lidar or radar data. TransLPC is able to remedy these issues: The structure of the transformer model is modified to allow for larger input sequence lengths, which are sufficient for large point clouds. Besides this, we propose a novel query refinement technique to improve detection accuracy, while retaining a memory-friendly number of transformer decoder queries. The queries are repositioned between layers, moving them closer to the bounding box they are estimating, in an efficient manner. This simple technique has a significant effect on detection accuracy, which is evaluated on the challenging nuScenes dataset on real-world lidar data. Besides this, the proposed method is compatible with existing transformer-based solutions that require object detection, e.g. for joint multi-object tracking and detection, and enables them to be used in conjunction with large point clouds.

\end{abstract}

\section{INTRODUCTION} 
\thispagestyle{FirstPage}
In recent years, transformer models \cite{vaswani_attention_2017} have revolutionized the field of natural language processing \cite{devlin2018bert, brown2020language}. Soon, they were also applied in various different fields, such as computer vision (e.g. image recognition \cite{dosovitskiy_image_2020}, segmentation \cite{ye2019cross}) and game theory \cite{noever2020chess}. With attention between all input tokens, global relationships can be modeled, different from convolutional neural networks, which only incorporate local interactions in every layer. Depending on the application, an input token can be an embedded word or image cell, or a different kind of feature vector.

Transformers have also been applied to object detection, i.e. to estimating bounding box parameters for each object in a measurement, such as their location, extent and class. Such a transformer-based object detector opens many possibilities for downstream tasks, e.g. tracking, prediction or planning. Another advantage is that they alleviate the need for non-maximum suppression as a post-processing step. One example is the Detection Transformer (DETR) \cite{carion_end--end_2020} that can be applied to image data, and that serves as basis for a joint tracking and detection model \cite{meinhardt_trackformer:_2021}. In \cite{meinhardt_trackformer:_2021}, the feature vectors as computed by DETR are propagated through time, with elegant track management solved implicitly by the transformer. Such a model would also be desirable for point cloud data, which is solved by 3DETR \cite{misra_end--end_2021}. However, existing transformer-based detection approaches are not suitable for large-scale point cloud datasets, i.e. those that span a large area per frame and can contain many measured points. This is caused by memory constraints, as the size of the attention matrix in a transformer grows quadratically with the input length. Such large datasets are common in the context of autonomous driving, where a lidar sensor can measure points in areas of $100 \times 100$ meters and beyond. One contribution of this work is that we investigate how to modify the general concept of DETR for large point clouds.

\begin{figure}[t]
	\centering
	\includegraphics[width=\columnwidth]{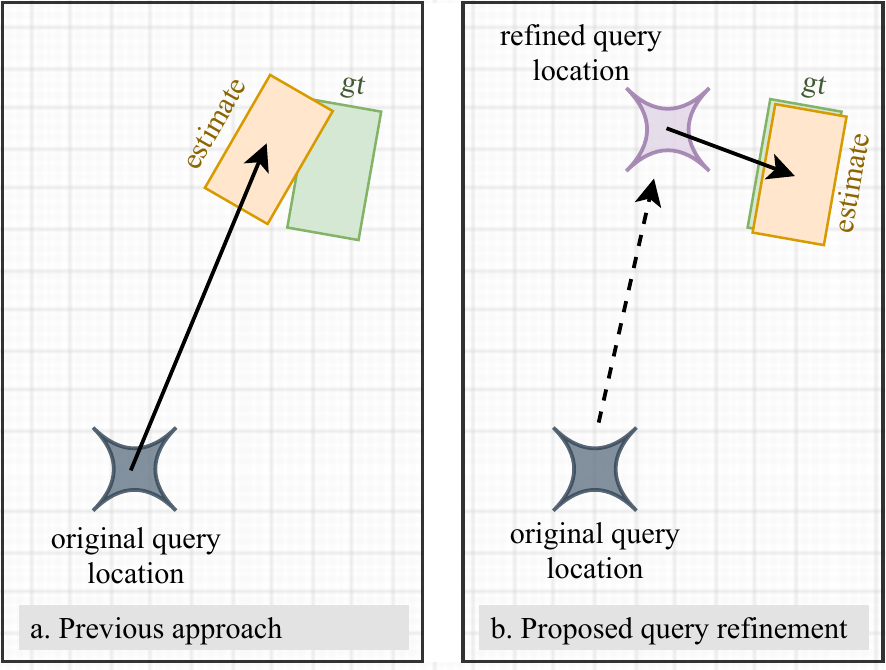}
	
	\caption{Illustration of query refinement. \textit{a.} Without refinement, a transformer decoder query might need to travel a large distance to estimate a bounding box, leading to inaccurate detections. \textit{b.} The original query is updated with its refined version, allowing for a more precise detection.}	\label{fig_query_refinement}
\end{figure}
Besides this, we shift our focus to the query vectors that are input to the transformer decoder model and that serve as slots for detection. In our model, each of them is assigned to a certain location with a positional encoding. We analyze the interpretability of these vectors as location-bound feature vectors and outline their potential for downstream tasks. Our analysis leads to the design of a novel query refinement technique that improves the detection accuracy on a large surface area, while retaining reasonable memory consumption, which is illustrated in Figure \ref{fig_query_refinement}. Rather than estimating a bounding box at a large distance from the query location directly (left), the refinement enables queries to travel iteratively towards the bounding box, allowing for a more accurate estimation.

To summarize, the main contributions of this paper are:
\begin{itemize}
	\item We present TransLPC, a novel transformer-based object detector that is applicable to large point clouds.
	\item A novel query refinement method is presented, which improves the detection accuracy, while causing negligible overhead. The method is thoroughly analyzed.
	\item Our model can serve as a detection platform for using existing transformer methods with large point clouds, such as joint detection and tracking \cite{meinhardt_trackformer:_2021}, which was previously hindered by memory constraints. This is explored in our follow-up work \cite{ruppel2022trans}.
\end{itemize}

\section{RELATED WORK}
In this section, related work that our method is built on is introduced. We briefly describe the transformer model \cite{vaswani_attention_2017}, as well as multiple existing methods for object detection.
\subsection{Transformers}\label{sec_transformers}
Components of the transformer model \cite{vaswani_attention_2017} that are relevant to this work are summarized in the following. A transformer consists of a decoder and an encoder, both of which contain attention modules. In the encoder, context information within the data is meant to be encoded. For this, self-attention between the input tokens $\bm{x}_i$, $i=1,\dots,N$ is employed, i.e. each token can interact with every other token, which results in the aforementioned quadratic memory requirement w.r.t. $N$. In the self-attention module, three vectors are computed from each token: A query, key and value vector. Each query is meant to query information from the other tokens, by comparing it to all other keys, thereby obtaining the attention weights. The attention result is then computed by weighting the values according to the attention weights.

The transformer decoder receives input tokens as well, which are different from the encoder input tokens and can have a different sequence length $M$. It employs self-attention between its tokens, as well as cross-attention. During the latter, the decoder input constitutes the query vectors, while keys and values stem from the encoder's output. Therefore, cross-attention enables the decoder to query information from the encoded input.
Both in the transformer decoder and the encoder, multiple layers are stacked on top of one another, each containing the aforementioned attention modules.
\subsection{Object Detection}
Object detection is a large research field with many different applications. Existing methods can be clustered based on their input modality and their output space. In the following, we focus on object detection in point clouds with 3D bounding boxes as output, which can roughly be categorized in three different approaches \cite{guo2020deep}: Even with point cloud input, \textit{grid-based} methods operate in 2D space, by projecting the input into a bird's eye view perspective first, e.g. \cite{simon2018complex, yu2017vehicle}, as well as
PointPillars \cite{lang_pointpillars:_2019}, or into a different kind of 2D space, such as through cylindrical projection \cite{li2016vehicle}. This allows them to use standard 2D convolutions, while the information from the missing dimension is encoded in feature vectors. Another approach is to divide the space into discrete volumes (\textit{voxels}) \cite{li20173d, engelcke2017vote3deep}, and to group points within the same voxel together. Many voxels can be empty due to the sparseness of the input data \cite{arnold2019survey}. Besides this, \textit{point-based} methods do not group the input into a fixed grid \cite{qi2017pointnet, qi_pointnet++_2017}. However, they do aggregate neighboring points, e.g. with ball query operations. Each of the aforementioned methods have in common that they compute feature vectors, which are assigned to a certain location or volume, whether it is on a regular grid or around a certain point. This is the definition of a backbone as it can be used with the model we propose in this work, allowing for a versatile choice of backbone as well as to benefit from the ongoing research in this area.

Transformer-based object detection is a relatively new development, based on which our method is built. DETR \cite{carion_end--end_2020} is a pioneering work in this field. The authors perform 2D object detection with a full transformer model, as well as a pretrained image backbone as prepocessing step. Learnt queries are input to the decoder, which are denoted object queries. These queries are transformed through the model, and bounding box estimates are obtained with an estimation head after the last layer. The DETR model has been extended in numerous works, such as \cite{zhu2020deformable, gao2021fast, meng2021conditional}. In \cite{misra_end--end_2021}, a similar method is applied to point cloud data from the indoor 3D dataset ScanNetV2 \cite{dai2017scannet}. Their model also utilizes a full transformer model with encoder and decoder, but no backbone for preprocessing. In their work, the bounding boxes are estimated relative to a reference location rather than absolute. Note that such indoor 3D datasets span a much smaller area than the large point clouds that are the focus of this work. There has also been research on transformer-based detection backbones, \cite{mao2021voxel} with a mixture of dilated and local attention and \cite{pan20213d} with blocks of local and global attention.
\section{PROPOSED MODEL}
\begin{figure*}[!t]
	\includegraphics[width=\textwidth]{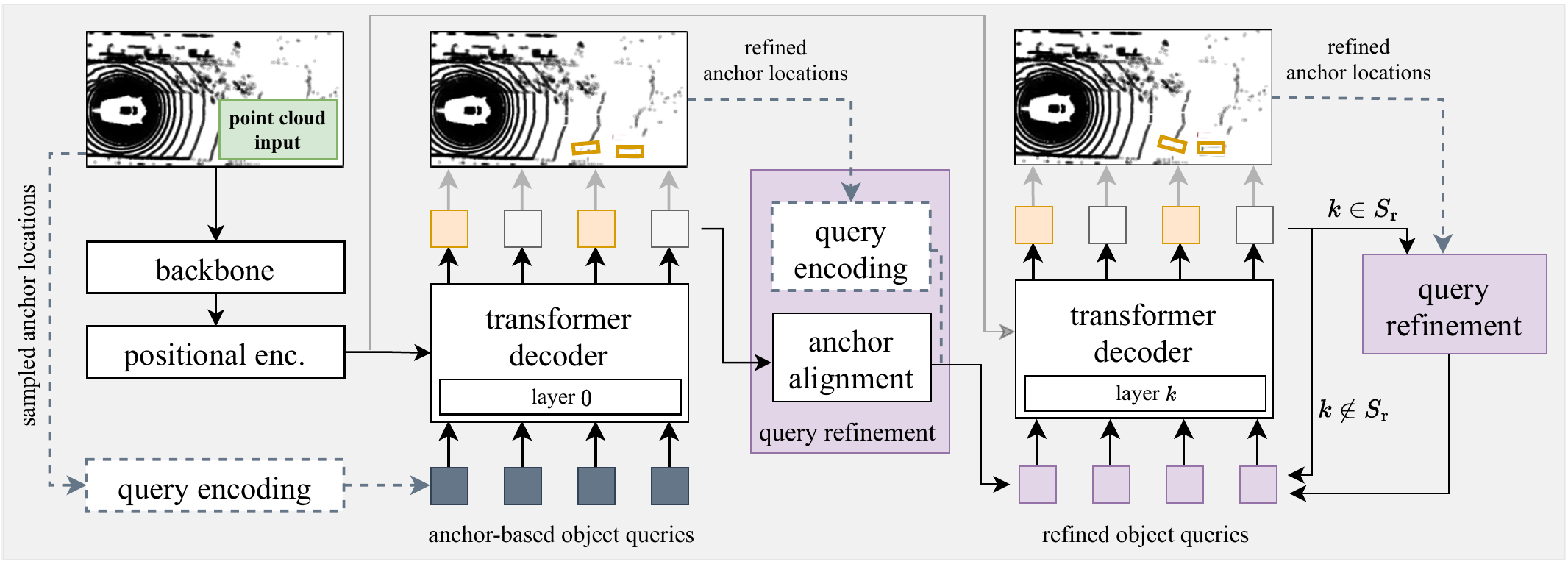}
	\caption{Detection model overview. The point cloud input frame is processed in a backbone and a positional encoding is applied to each feature vector. For the first transformer decoder layer, anchor-based object queries are generated by encoding sampled locations from the data. The result of this layer is enhanced by query refinement before it is passed to the next (purple). Subsequent layers can either receive the propagated result of the previous one, or another refinement step is added.}
	\label{fig_detector2}
\end{figure*}

An overview of TransLPC, the proposed transformer detector for large point clouds can be found in Fig. \ref{fig_detector2}. The model consists, among other things, of a backbone of choice, a transformer decoder and a novel query refinement module. Each of its components will be introduced in the following.
\subsection{Allowing for a Large Input Length}
As mentioned above, one fundamental limitation of the standard transformer model \cite{vaswani_attention_2017} is that the memory requirement of its attention modules grows quadratically with the number of input tokens $N$. Due to this circumstance, the transformer in DETR \cite{carion_end--end_2020} operates on a grid of only $45 \times 35$ cells, i.e. $N=1575$ tokens. In \cite{misra_end--end_2021}, a set aggregation downsampling \cite{qi_pointnet++_2017} is used on the input point cloud to reduce the input to the transformer model to $N=2048$. In numerous experiments, we found that such aggressive downsampling is not feasible for point clouds that span a large area, which are the focus of this work. Both a point based downsampling, such as \cite{qi_pointnet++_2017}, as well as a birds-eye-view grid based downsampling \cite{lang_pointpillars:_2019} result in tokens belonging to so broad influence regions or so large grid cells that it disrupts the detection quality. There is no clear dividing line concerning what is regarded as a large point cloud. In the context of this work, we consider a point cloud to be large if the aforementioned downsampling is not feasible, since a desired detection resolution can not be reached with it.

One option to handle such a point cloud may be to utilize one of the modified transformer models that aim to mitigate the quadratic memory requirement \cite{wang_linformer:_2020} \cite{tay_efficient_2020}. Many models achieve linear complexity by approximating the transformer's attention module. However, the input sequence length is still limited, while the attention mechanism is only an approximation of the original design. These two aspects explain why we found what functions better for detection on large point clouds: Leaving out the transformer encoder entirely. We find that a backbone, such as PointPillars \cite{lang_pointpillars:_2019}, can sufficiently fulfill the task of encoding context information. With this removal, the transformer looses some of its global context. However, in a bird's eye view perspective of a large point cloud, it is reasonable to encode mainly local information rather than the context many meters away. The transformer \textit{decoder} on the other hand, which is still part of the proposed model, retains its global context through cross-attention.

The backbone output is flattened to a sequence of tokens (feature vectors) of length $N$ and a positional encoding as in \cite{carion_end--end_2020} is applied. As mentioned in the introduction, many different backbones could be used, as long as they supply feature vectors that each belong to a certain location, whether on a grid or in 3D space. Since the transformer encoder bottleneck has been removed, a much larger length $N$ is possible than before, ca. by a factor of $20$, depending on the number of decoder input queries $M$. The largest attention matrix of the model is now located in the decoder's cross-attention module and it is of size $N \times M$, where $N \gg M$.
\subsection{Transformer Architecture}
The flattened backbone output with positional encoding serves as input to a standard transformer decoder \cite{vaswani_attention_2017}, constituting the keys and values in cross-attention. The decoder has $K$ layers and the query, key and value vectors are of dimension $d$.

Following \cite{carion_end--end_2020}, we formulate object detection as a set prediction problem, where the $M$ transformer decoder queries, denoted as object queries, serve as a fixed number of slots for possible objects. They can communicate among one another with self-attention as well as query information from the input data during cross-attention.

Inspired by \cite{misra_end--end_2021}, we generate the object queries by first using a farthest point sampling on the input point cloud to obtain locations $\bm{\rho}_i$, with $i=1,\dots,M$. Then, these locations are encoded with a Fourier encoding \cite{tancik_fourier_2020}:
\begin{equation}\label{eq_query_encoding}
\bm{y}_i=\textrm{FFN}\left[\sin(\bm{B}\bm{\rho}_i), \cos(\bm{B}\bm{\rho}_i)\right],
\end{equation}
where $\bm{B}\in \mathbb{R}^{\frac{d}{2}\times3}$ is a matrix filled with entries drawn from a normal distribution. The feed-forward network (FFN) is trained with the rest of the model and consists of two layers with ReLU activation, and $\bm{Y}_0=\{\bm{y}_i\}_{i=1}^M$ denotes the set of tokens that is input to the first decoder layer, i.e. layer $0$.

We name the locations $\bm{\rho}_i$ \textit{anchor locations}, since they serve as a prior location for object detection, encouraging each query to search for objects in a certain area. However, the query is not restricted from estimating a bounding box at a significant distance from its anchor location, if necessary. Note that the anchor locations are different from anchor boxes, which are commonly used in other detection approaches, such as \cite{lang_pointpillars:_2019}.

For point cloud data, location-bound object queries are superior to learnt queries \cite{misra_end--end_2021}, because the data is inherently sparse. As shown in \cite{carion_end--end_2020}, learnt queries do learn to primarily attend to a certain area in an image. With point cloud data, however, they would also need to waste resources on finding locations that actually contain data within their area of interest.

Even before the object queries $\bm{y}_i$ are input to the transformer decoder for the first time, they are interpretable as feature vectors belonging to a (prior) bounding box. We can apply an estimation head to each of them to obtain the following parameters:
\begin{equation}\label{eq_box_params}
\bm{b}_{[\bm{y}_i]} = (\Delta x,\Delta y,\Delta z,w,l,h, \sin(\gamma), \cos(\gamma), v_x,v_y, {\tt cls}),
\end{equation}
where $(x, y,z)=(\Delta x,\Delta y,\Delta z)+\bm{\rho}_i$ is the box position estimated relative to the corresponding anchor location $\bm{\rho}_i$, 
$(w,l,h)$ are the width, length and height, $\gamma$ denotes the orientation, $(v_x,v_y)$ the velocities in $x$ and $y$ directions and ${\tt cls}$ is a class identifier. The same box parameter estimation head can be applied again after each successive decoder layer, giving insights into the movement of each query from its first anchor location until its final box estimate.

\subsection{Query Refinement}
\begin{figure*}[!t]
	\centering%
	\subfloat[Propagation approach]{
		\captionsetup[subfigure]{labelformat=empty}
		\makebox[\textwidth]{
	\subfloat[Layer 0]{\includegraphics[trim={7cm 8cm 6.5cm 1.7cm},clip,width=0.47\columnwidth]{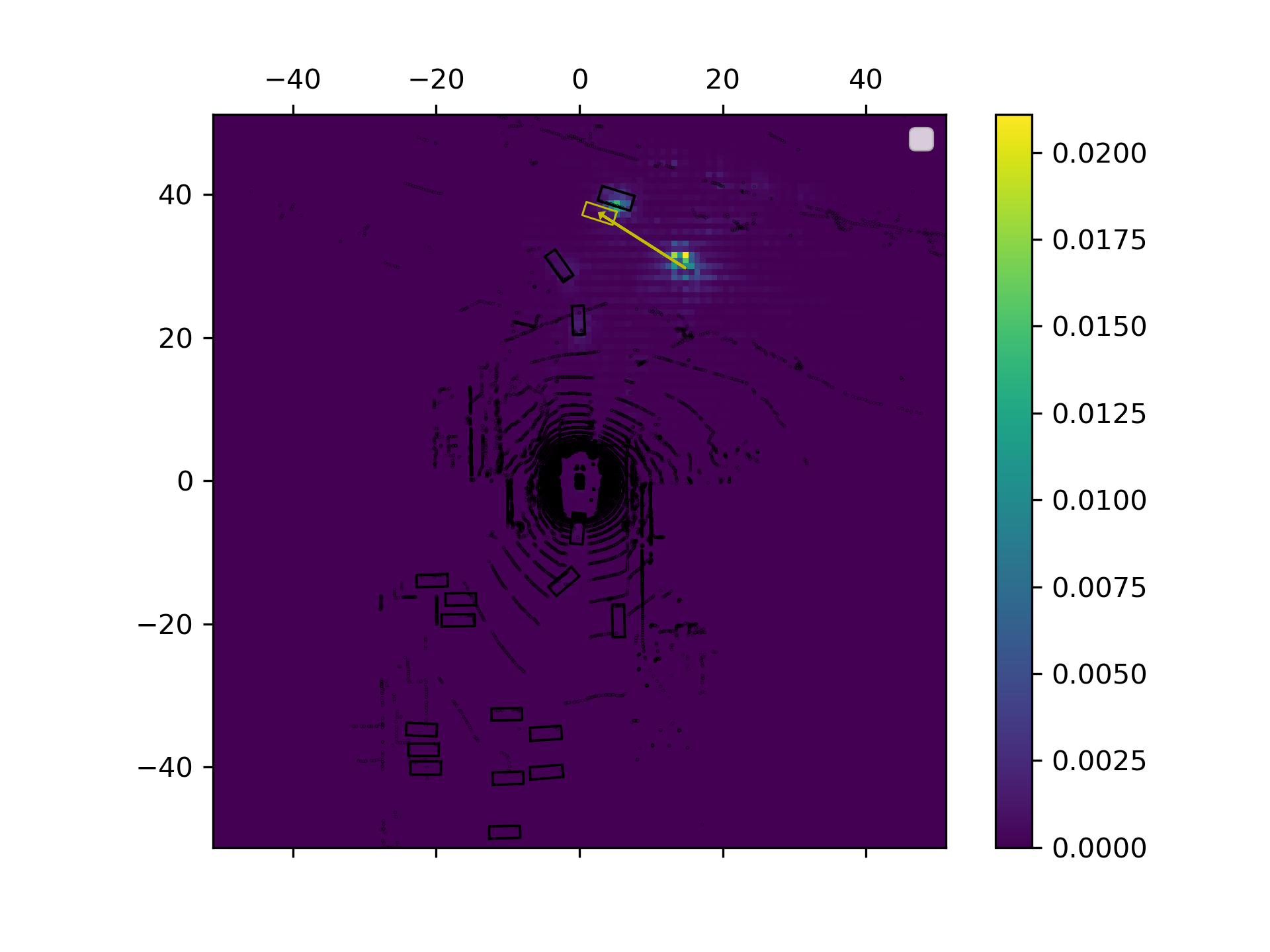}}
	\hfill
	\subfloat[Layer 1]{\includegraphics[trim={7cm 8cm 6.5cm 1.7cm},clip,width=0.47\columnwidth]{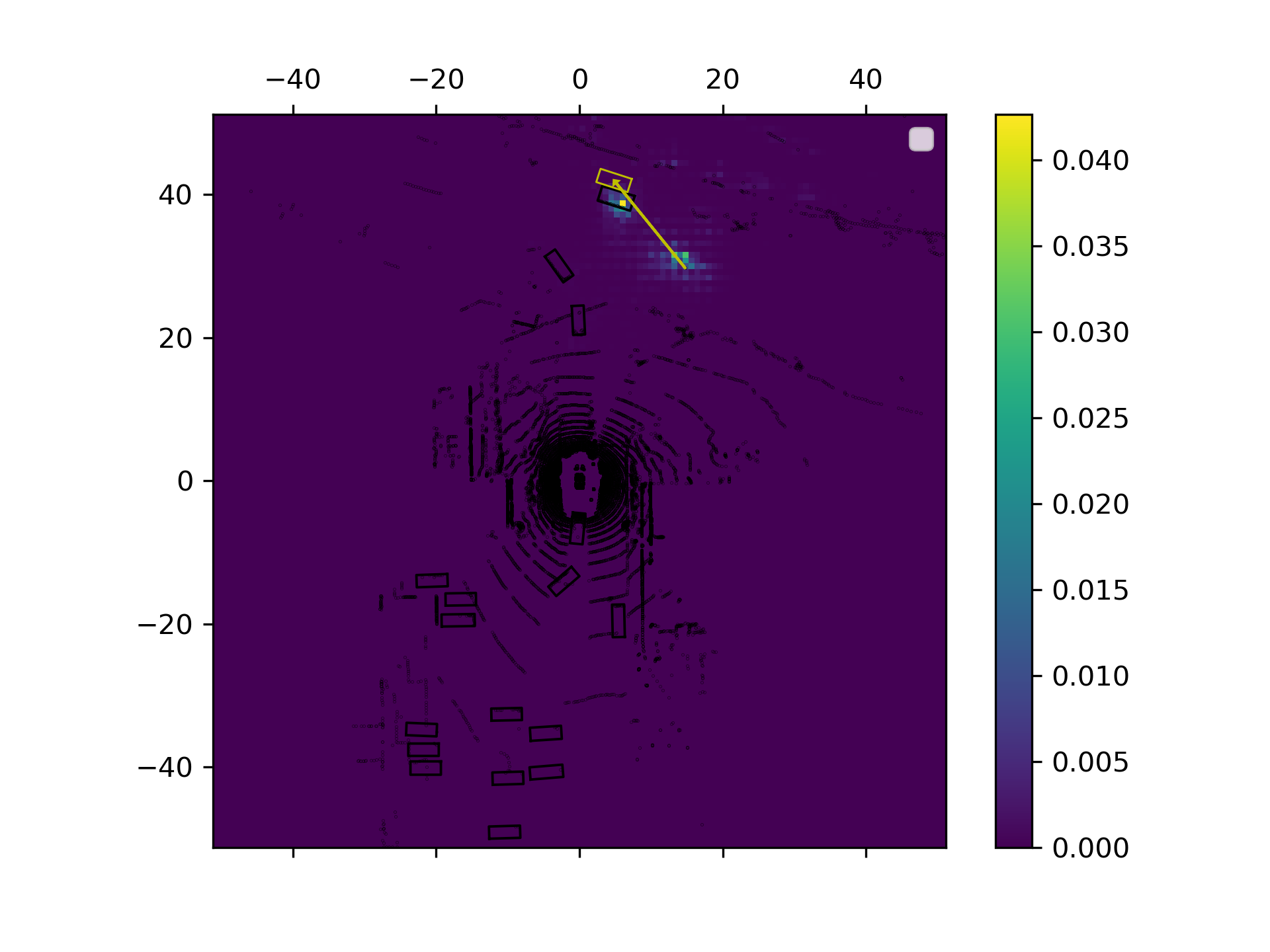}}
	\hfill
	\subfloat[Layer 4]{\includegraphics[trim={7cm 8cm 6.5cm 1.7cm},clip,width=0.47\columnwidth]{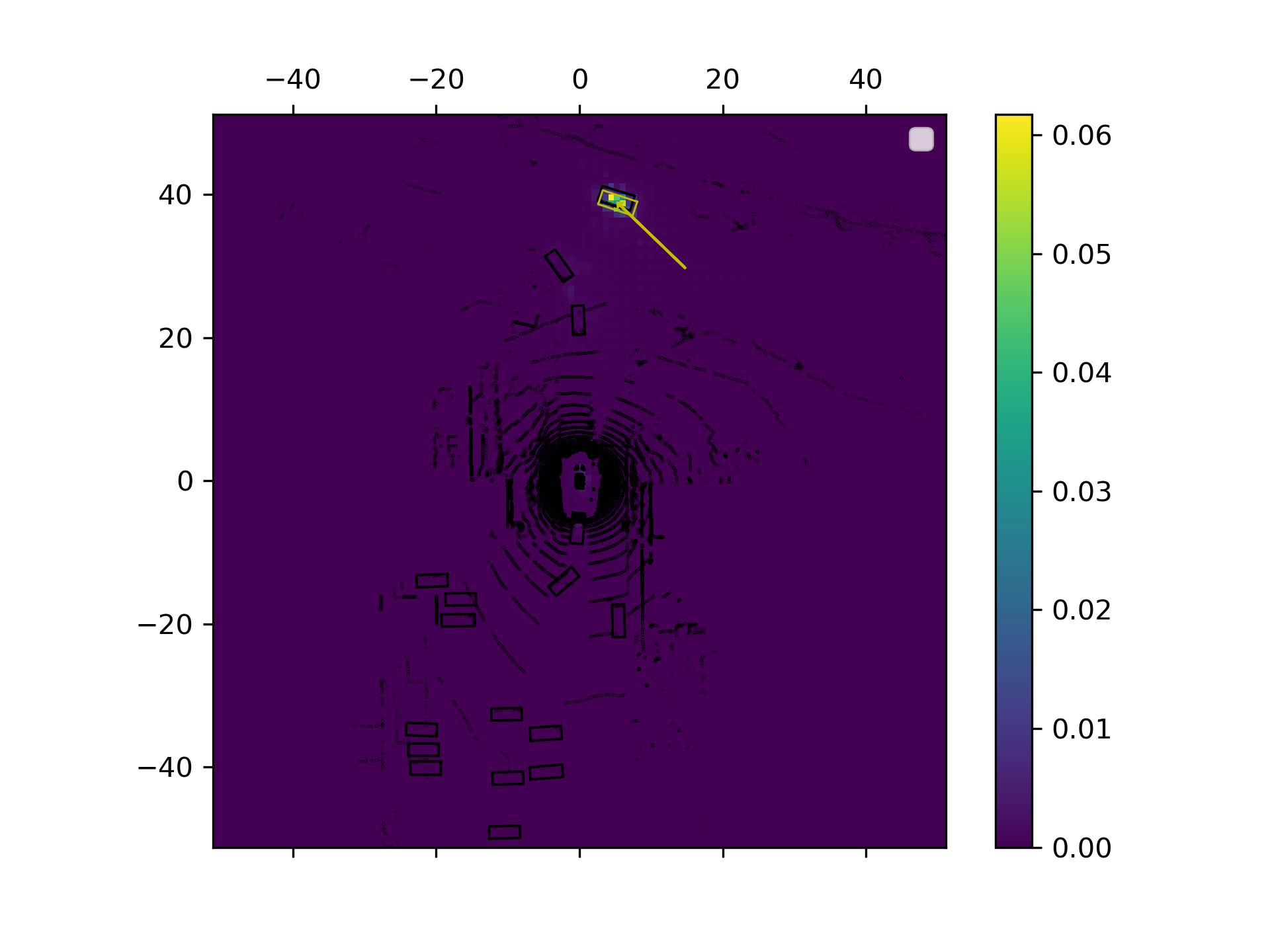}}
	\hfill
	\subfloat[Layer 5]{\includegraphics[trim={7cm 8cm 6.5cm 1.7cm},clip,width=0.47\columnwidth]{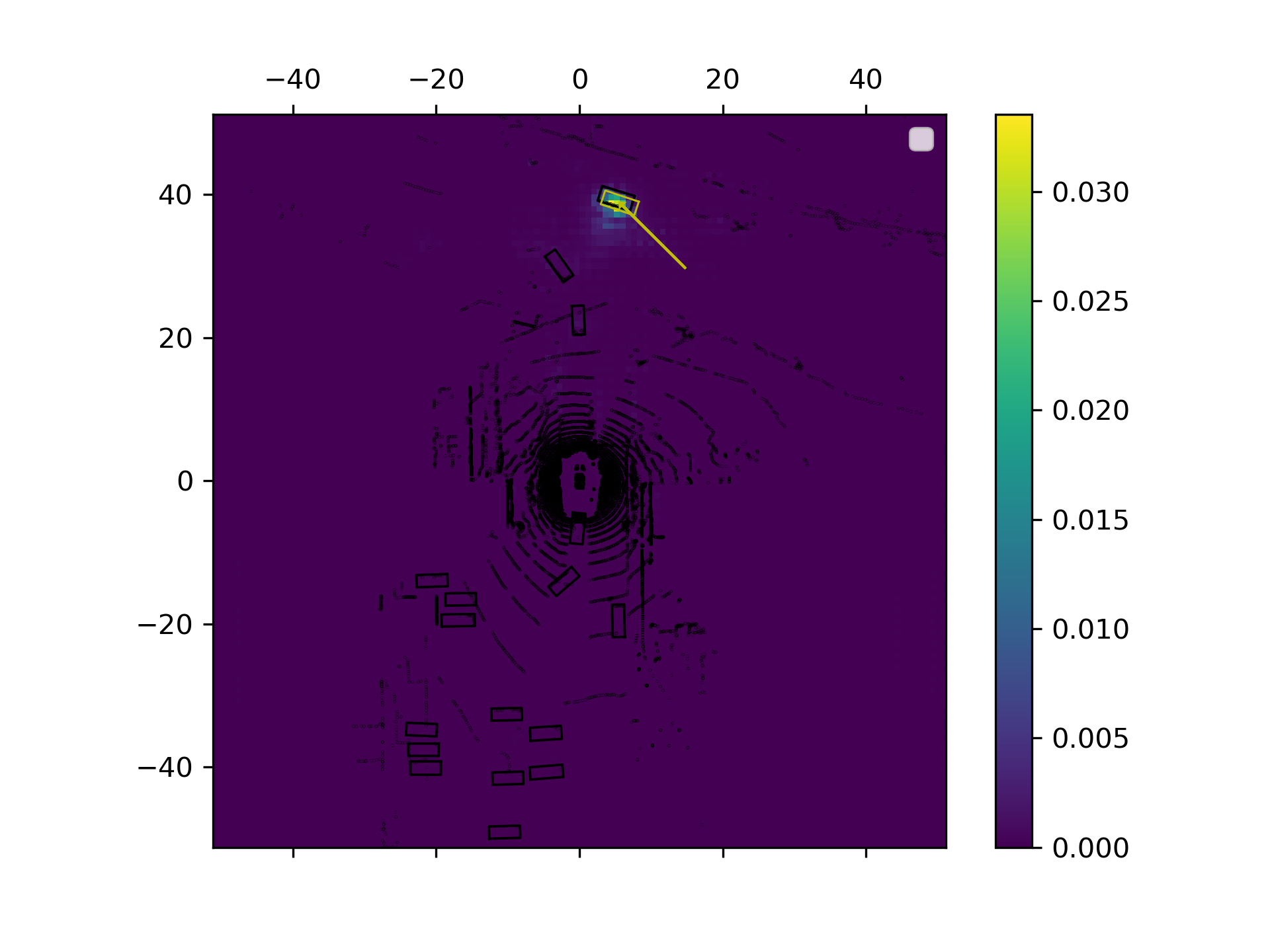}}\addtocounter{subfigure}{-4}}}\\
\addtocounter{subfigure}{-4}
	\subfloat[Query refinement approach]{\captionsetup[subfigure]{labelformat=empty}
		\makebox[\textwidth]{
		\subfloat[Layer 0]{\includegraphics[trim={7cm 8cm 6.5cm 1.7cm},clip,width=0.47\columnwidth]{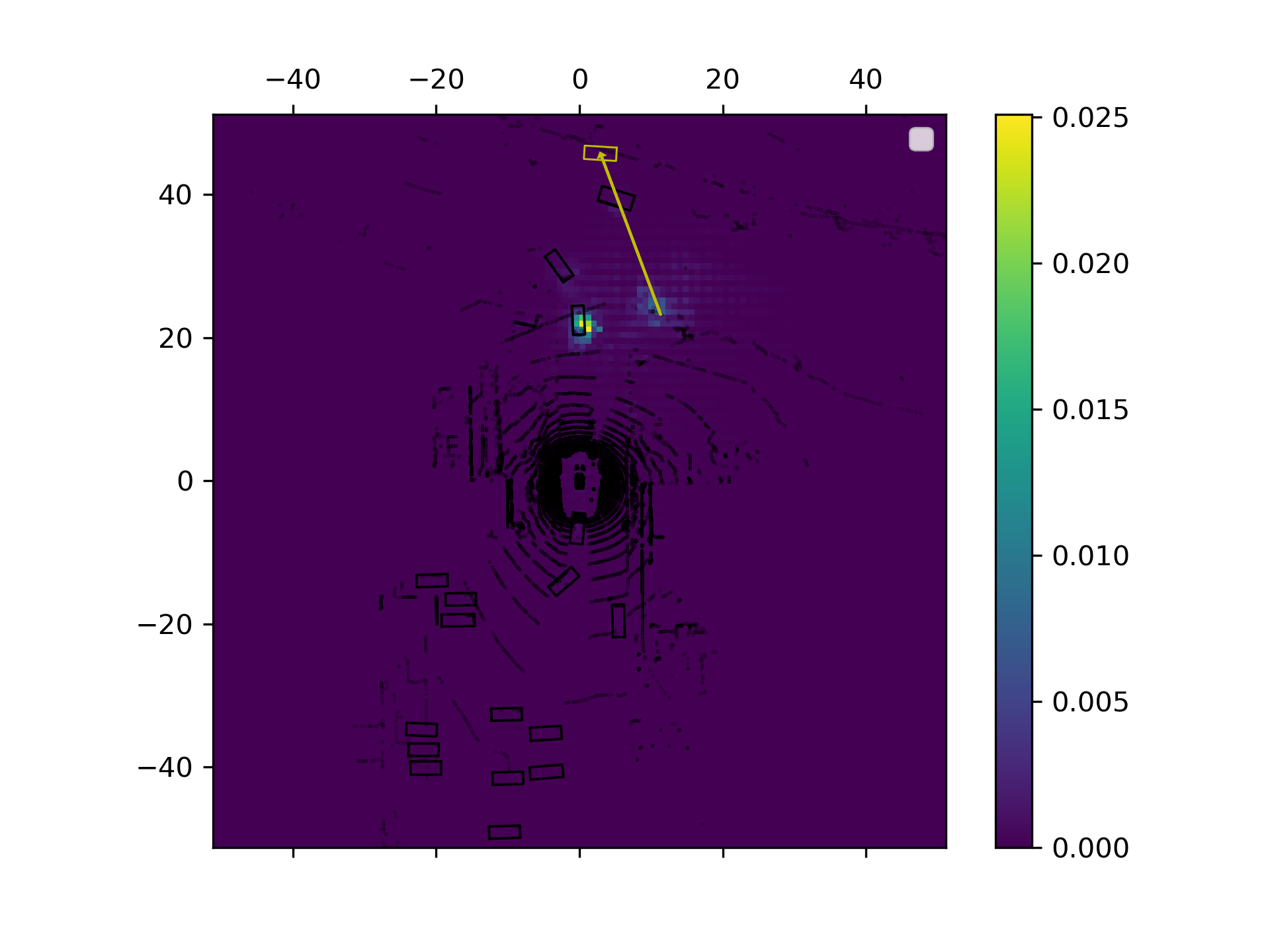}}
	\hfill
	\subfloat[Layer 1]{\includegraphics[trim={7cm 8cm 6.5cm 1.7cm},clip,width=0.47\columnwidth]{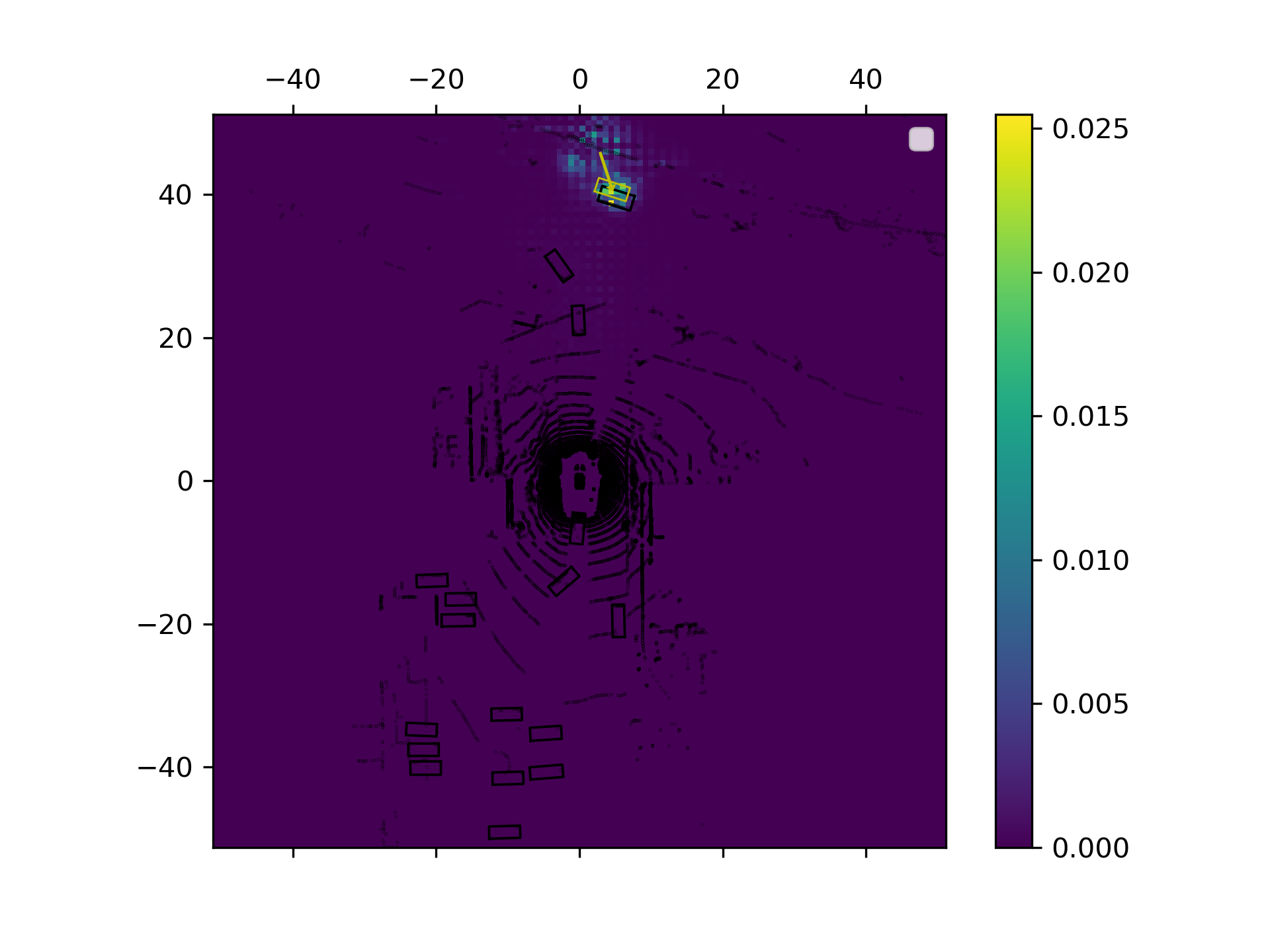}}
	\hfill
	\subfloat[Layer 4]{\includegraphics[trim={7cm 8cm 6.5cm 1.7cm},clip,width=0.47\columnwidth]{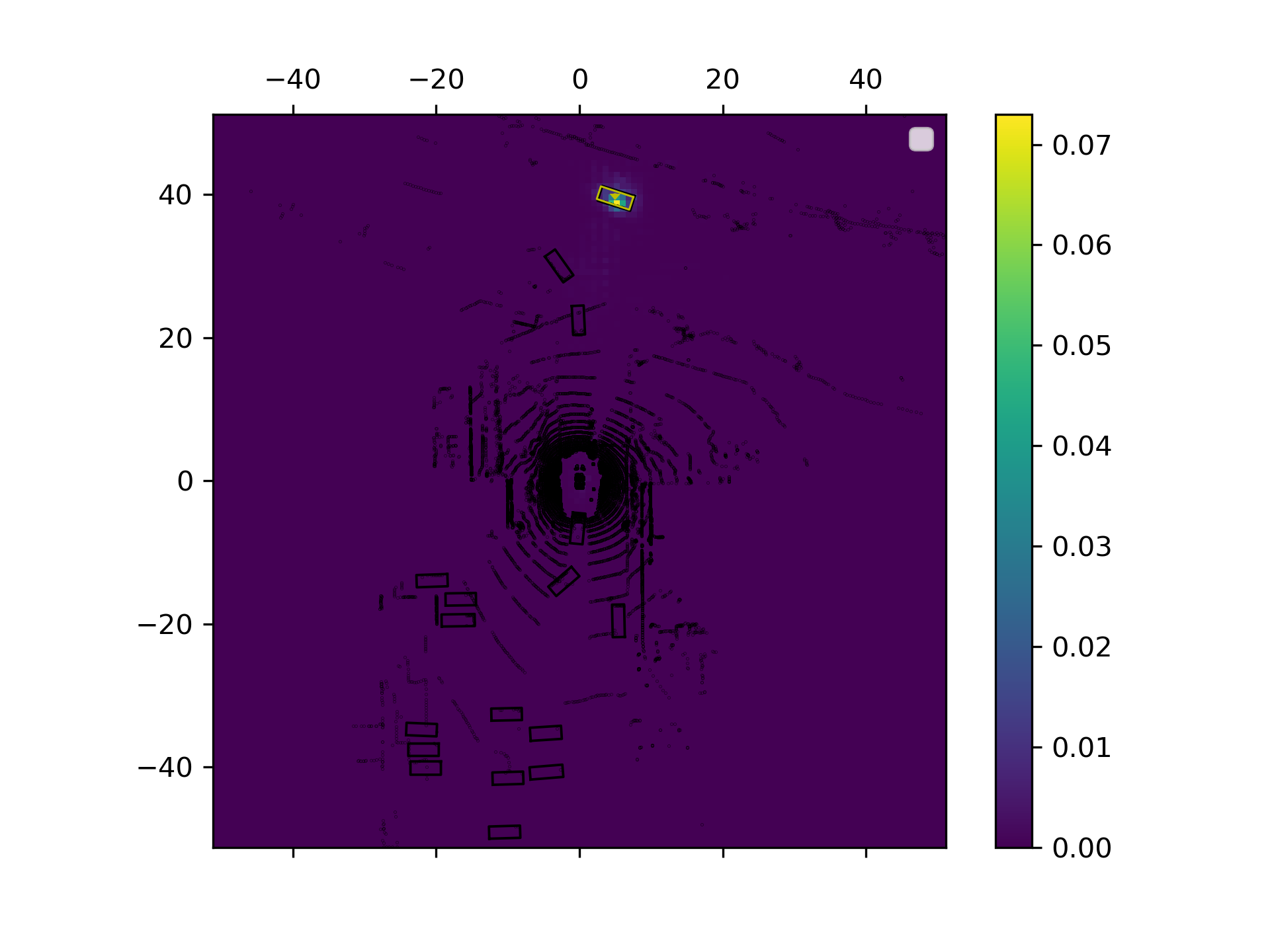}}
	\hfill
	\subfloat[Layer 5]{\includegraphics[trim={7cm 8cm 6.5cm 1.7cm},clip,width=0.47\columnwidth]{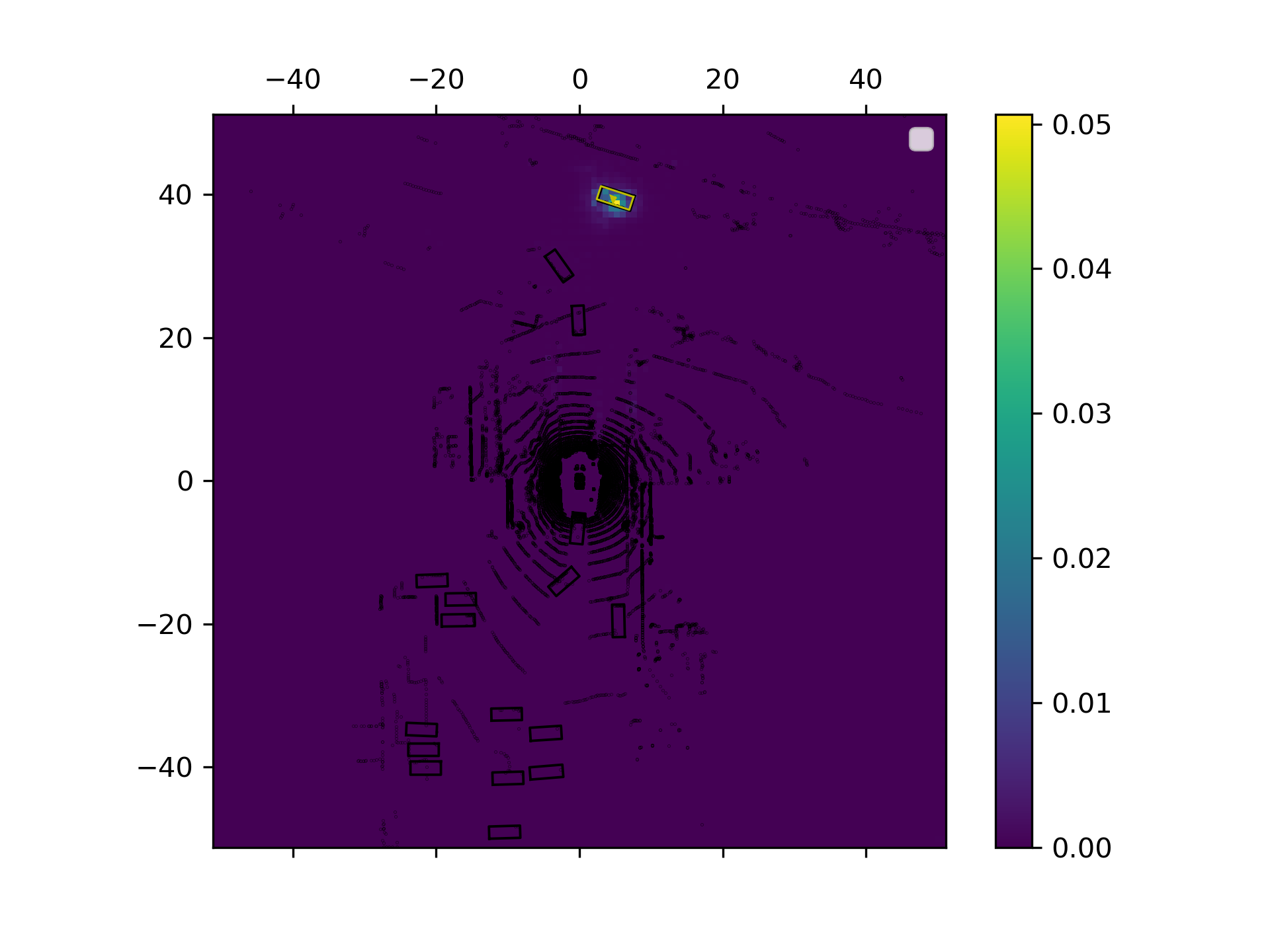}}}}
	\caption{Visualization of cross-attention weights in the transformer decoder layers for the \textit{propagation} approach (top) and \textit{query refinement} approach (bottom). An excerpt of the full bird's-eye view grid is pictured. The (square) input grid cells are colored according to their attention weight towards one selected object query, where brighter colors denote higher values. Yellow arrows illustrate the movement between the latest query anchor location and the bounding box (yellow box) that is computed from it after the given layer. In black, GT boxes are displayed, as well as the lidar input data points. Without refinement (top), the query is forced to estimate its box from a distance, leading to an inaccurate detection.}
	\label{fig_attn}
	\vspace{-10pt}
\end{figure*}

In a standard transformer decoder, a subsequent layer receives the previous layer's output as its input, which we denote as \textit{propagation approach} in the following. A positional encoding, such as the anchor location encoding, can be supplied to each layer in addition, which is also part of this method, following \cite{carion_end--end_2020, misra_end--end_2021}. We found, however, that this approach leaves room for improvement. As illustrated in Figure \ref{fig_query_refinement} (left), if a query needs to travel a large distance towards its estimated bounding box, this can result in a less accurate estimation than if the query was closer to the target. This phenomenon is observable quantitatively, which can be found in Figure \ref{fig_rmse_query_movement_propagation_approach} (top). It is also remarkable that with the \textit{propagation} approach, in a fully trained model, the first layer seems to be essential in deciding the area of interest of each query, i.e. the later layers mostly refine the initial bounding box guess. One example for this behavior is plotted in Figure~\ref{fig_attn} (top row). Here, the attention weights between one object query and the input data grid is pictured for subsequent decoder layers. The movement between the query anchor location and the box that was estimated from its layer's output through the estimation head (yellow box) is pictured as yellow arrow. The model is forced to refine this box at a distance, since the original query anchor location remains unchanged through the layers.

\begin{figure}[t]
	\centering
	\includegraphics[width=\columnwidth]{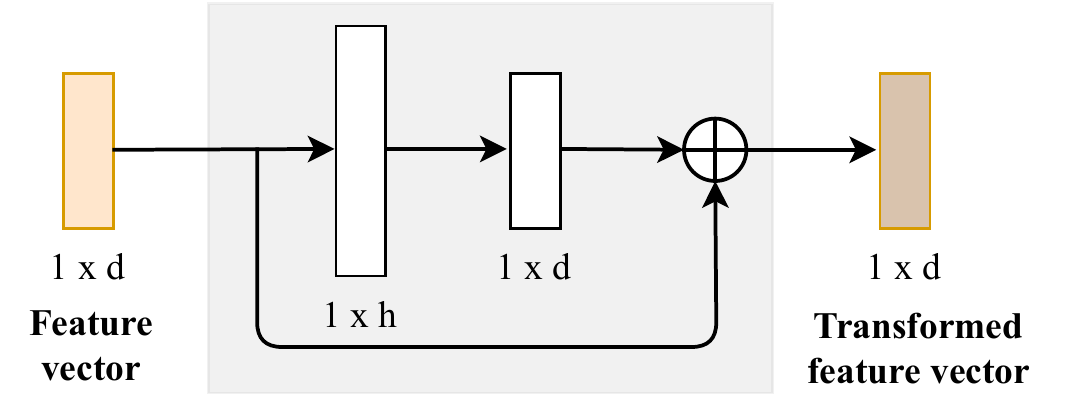}
	\caption{Anchor alignment module overview.}	\label{fig_AAM}
	\vspace{-12pt}
\end{figure}

These observations lead to the design of the proposed \textit{query refinement} approach: After the first decoder layer (layer $0$), the box parameter estimation head is applied to all layer output tokens $\bm{z}_i^{(0)}$, obtaining $M$ bounding box estimates $\bm{b}_{[\bm{z}_i^{(0)}]}$. This first guess now constitutes new, refined anchor locations
\begin{equation}
	\bm{\rho}_i^{(1)} = (\Delta x,\Delta y,\Delta z)+\bm{\rho}_i^{(0)},
\end{equation}
$i=1,\dots, M$ for layer $1$, which are encoded as in Eq. \ref{eq_query_encoding} to obtain vectors $\bm{y}_i^{(1)}$. After refinement, a query estimates its bounding box relative to its new, refined anchor location, which is generally closer to the object it is aiming to detect. To be able to retain the information that is contained within the output of layer $0$, i.e. $\bm{z}_i^{(0)}$, (besides the location that was computed from it), we propose to transform the vectors to align with their new anchor locations:
\begin{equation}\label{eq_anchor_align}
	\bm{\tilde{z}}_i^{(0)}=\textrm{AAM}\left(\bm{z}_i^{(0)}\right),
\end{equation}
where AAM denotes the proposed anchor alignment module that consists of two FFN layers with ReLU activation and a skip connection from the input to the output. Its light-weight structure is pictured in Fig. \ref{fig_AAM}. Note that, before refinement, the parameters $(\Delta x,\Delta y,\Delta z)$ can be large, i.e. the layer output token $\bm{z}_i^{(0)}$ can constitute a box that is far away from its anchor $\bm{\rho}_i^{(0)}$. After refinement, the input to the subsequent layer has a new anchor $\bm{\rho}_i^{(1)}$ that is closer to the object it is aiming to detect and it is aligned with its new anchor through the AAM.

The set of queries that is input to the following layer $1$ of the decoder is $\bm{Y}_1=\{\bm{\tilde{z}}_i^{(0)}+\bm{y}_i^{(1)}\}_{i=1}^M$. Depending on the task, it can be useful to repeat the query refinement for later decoder layers:
\begin{equation}
	\bm{Y}_k= \begin{cases}
					\{\bm{\tilde{z}}_i^{(k-1)}+\bm{y}_i^{(k)}\}_{i=1}^M& \textrm{for } k\in S_\textrm{r} \\
					\{\bm{z}_i^{(k-1)}+\bm{y}_i^{(j)}\}_{i=1}^M        & \textrm{for } k\not\in S_\textrm{r} \textrm{ and } k\neq0\\
					\{\bm{y}_i^{(0)}\}_{i=1}^M & \textrm{for }k=0,
			\end{cases}
\end{equation}
where $S_\textrm{r}$ contains the indices of those layers before which the query refinement is applied and $j = \max \{l\mid l<k\land l\in S_\textrm{r}\}$ enforces that the latest anchor location encoding is always added to the propagated query.

In Figure \ref{fig_attn} (bottom), query refinement was performed after every layer, resulting in a very small distance between the last query anchor location and its estimated box, which allows for a more accurate detection. Besides this, it is observable that the query is able to iteratively travel towards its box.

The query refinement step causes negligible overhead, since the AAM is a small module with only two layers and besides the AAM, the same anchor location encoding is used as the one before the first layer, i.e. no additional parameters are needed.
\subsection{Training and Inference}\label{training_details}
TransLPC is trained with auxiliary losses after every decoder layer. To obtain the loss, the box estimation head is used to compute $M$ sets of bounding box parameters from each layer's output. A Hungarian algorithm matches every ground truth (GT) box to the closest box estimate. Since $M$ is generally larger than the number of GT objects, the remaining estimates are matched to the 'no-object' class, discouraging duplicate detections, following \cite{carion_end--end_2020}. An $\ell_1$-loss is applied to the difference between estimated and GT boxes. Besides this, an auxiliary loss is employed before the first decoder layer, aligning the queries with their first anchors. As backbone, a pretrained PointPillars \cite{lang_pointpillars:_2019} backbone is used, which outputs a grid of feature vectors, and its weights are refined during TransLPC training. The grid is flattened to a sequence of feature vectors, while the positional encoding is computed from the grid locations.

The anchor alignment module is trained separately from the rest of the model and is used with fixed weights afterwards. To obtain AAM input data and ground truth, a fully trained variation of TransLPC, which follows the $propagation$ approach, is used. It computes decoder output tokens $\bm{z}_i$ from the point cloud, which are input to the AAM. Now, ground truth data is obtained with the box parameter estimation head: Both the vectors $\bm{z}_i$ and $\textrm{AAM}\left(\bm{z}_i\right)$ are input to this head. All box parameters shall be unchanged by the AAM module, except for the location deltas $(\Delta x,\Delta y,\Delta z)$, which are supposed to be zero, in order to align with their new anchor position.

During inference, the box parameter estimation head is applied to the last layer's output of TransLPC, obtaining box estimates relative to the latest refined query anchor locations. A box is considered valid if it is not assigned to the 'no-object' class.

\subsection{Expansion to Downstream Tasks}
The transformer decoder queries as presented in this paper are location-bound vectors that contain successively more information about an object as they are transformed through the decoder layers and by query refinement. This makes them valuable for downstream tasks, as they encode the object's information in latent space rather than only in the form of low dimensional box parameters. Such vectors can be propagated in time for tracking, as introduced in \cite{meinhardt_trackformer:_2021}, where the multi-object tracking task is solved implicitly with a transformer. TransLPC is compatible with such a model, which we explore in \cite{ruppel2022trans}.
\section{RESULTS}
We evaluate TransLPC with different kinds of query refinement placements and compare it to the aforementioned \textit{propagation} approach, both in terms of known metrics as well as in a novel analysis, which examines the movement between an original query location and the estimated bounding box center. The results are detailed in the following.
\subsection{Detection results}
TransLPC is trained on the nuScenes \cite{caesar_nuscenes:_2020} {\tt train} dataset (lidar data) with a batch size of $30$ on an NVIDIA A100 GPU. We set $K=6$, $d=256$, $N=16384$ and $M~=~100$. First, the model is trained for $300$ epochs with no query refinement until it reaches saturation. Then, the anchor alignment module is trained for 30 epochs. Afterwards, the TransLPC training is continued with query refinement and the trained AAM, for an additional $300$ epochs. The learning rate is initialized at $l_r=10^{-4}$ and it is decreased by a factor of $10$ every $200$ epochs. The evaluation is performed on the nuScenes {\tt val} dataset and the nuScenes metrics are presented, which are defined in \cite{caesar_nuscenes:_2020}.

In Table \ref{tab:object_detection}, the results of the proposed model can be found, as well as those of the \textit{propagation} approach for comparison and of the PointPillars detection model \cite{lang_pointpillars:_2019}, which we consider our baseline. Note that the latter requires a non-maximum suppression, which TransLPC does not (see Section \ref{nms}). We find that the \textit{propagation} approach can already outperform the baseline slightly, while the additional \textit{query refinement} further increases detection accuracy significantly.

\begin{table}[h]
	\setlength\tabcolsep{0pt} 
	\caption{Object detection results (car)}
	\label{tab:object_detection}
	\begin{tabular*}{\columnwidth}{@{\extracolsep{\fill}} ll cccccc}
		\toprule
		Method  & AP$\uparrow$ & ATE (m)$\downarrow$& ASE (1-IOU) $\downarrow$& AOE (rad)$\downarrow$\\
		\midrule
		PointPillars \cite{lang_pointpillars:_2019} &0.684 &
		0.281& 0.164&0.204\\
		Propagation approach  &0.713&	0.310&	0.158&0.086 \\
		TransLPC (ours)   &\textbf{0.759}&\textbf{0.251}&\textbf{0.157}&\textbf{0.080} \\
		\bottomrule
		
	\end{tabular*}
\vspace{-4pt}
\end{table}
\subsection{Query Refinement}
Whether query refinement is applied or not makes a difference. First, we analyze the dependence of box location errors to the distance that the respective query traveled to its box. In Fig. \ref{fig_rmse_query_movement_propagation_approach} (top), a fully trained model with no query refinement is evaluated. Here, query travel lengths of up to $15$ meters are common, and even larger ones occasionally occur. A dependence of the median box location error to query travel length is clearly visible. In Fig. \ref{fig_rmse_query_movement_propagation_approach} (bottom), a model with query refinement after each layer is examined. In this case, two different kinds of query travel lengths can be considered. LQ denotes a plot w.r.t. the distance between the latest query location and the estimated bounding box. The histogram shows that most LQ travel lengths are smaller than $4$ meters due to refinement, while distances up to $8$ meters occur very rarely (in $0.245\%$ of observed boxes). For comparability with the \textit{propagation} method, the median location errors are plotted in terms of the distance that the first query anchor locations (FQ) traveled towards the final bounding boxes, i.e. the ones that were selected before layer~$0$. We observe that the slope of the error curve is flattened by the anchor refinement, clearly reducing the adverse effect of large query travel lengths that was observable before.

In Table \ref{tab:refinement_variants}, different variants of query refinement are compared in terms of the nuScenes metrics \cite{caesar_nuscenes:_2020}. We differentiate between refinement only after layer $0$ (\textit{Once}), after every second layer and after every layer. Besides this, \textit{propagation} denotes the method with no query refinement at all. We find that adding the query refinement once after layer $0$ achieves a significant improvement compared to \textit{propagation}. Adding more refinement steps later on can still improve the results, but not as significantly as the first refinement.

It can be concluded that query refinement is an effective method to improve accuracy at a very low cost concerning the small additional overhead it causes. The method combines the advantages of learnt queries \cite{carion_end--end_2020} with those of estimation relative to a reference location \cite{misra_end--end_2021}: A query can receive an initial guess through farthest point sampling based on the input data, rather than having to search for cells that actually contain measurements in sparse data. On the other hand it is flexible like learnt queries, being able to move to its region of interest.
\begin{figure}[t]
	\vspace{10pt}
\begin{tikzpicture}

\definecolor{color0}{rgb}{0.3803921568627451, 0.6627450980392157, 0.3607843137254902}
\definecolor{color1}{rgb}{1.0, 0.5725490196078431, 0.1411764705882353}

\begin{axis}[
height=151.86214363138544,
tick align=outside,
tick pos=left,
width=190.71811,
x grid style={white!69.0196078431373!black},
xmin=-1.8, xmax=24,
xlabel={Query travel length [m]},
ylabel={Location error [m]},
xtick style={color=black},
y grid style={white!69.0196078431373!black},
ylabel near ticks,
ymin=-2.51492022462189, ymax=56.7002523962408,
ytick style={color=black},
legend cell align={left},
legend style={at={(0.04,0.8)},anchor=south west,fill opacity=0.8, draw opacity=1, text opacity=1, draw=white!80!black}
]
\addplot [semithick, color0]
table {%
	2 0.419549256563187
	6 0.39963686466217
	10 0.638022720813751
	14 1.15661382675171
	18 5.2973895072937
	22 19.5405025482178
};
\addlegendentry{Location error}

\addplot+[gray,opacity=0.3, ybar interval,mark=no, fill=lightgray] 
table [x expr=\thisrow{X}, y expr=\thisrow{Y}*(1.4)]{%
	X Y
	0 29.8833332061768
	4 24.6200008392334
	8 8.55000019073486
	12 2.23666667938232
	16 0.529999971389771
	20 0.123333334922791
	24 0.0366666652262211
	28 0.0233333334326744
	32 0.0166666675359011
	36 0.00999999977648258
	40 0.00666666682809591
};
\addlegendentry{Histogram}
\path [draw=color0, semithick]
(axis cs:2,0.196285665035248)
--(axis cs:2,1.49478057026863);

\path [draw=color0, semithick]
(axis cs:6,0.176678530871868)
--(axis cs:6,1.36745424568653);

\path [draw=color0, semithick]
(axis cs:10,0.323243409395218)
--(axis cs:10,4.24038273096085);

\path [draw=color0, semithick]
(axis cs:14,0.660858139395714)
--(axis cs:14,15.5382318496704);

\path [draw=color0, semithick]
(axis cs:18,4.36999273300171)
--(axis cs:18,26.2786116600037);

\path [draw=color0, semithick]
(axis cs:22,16.2232275009155)
--(axis cs:22,54.0086536407471);

\path [draw=color0, semithick]
(axis cs:26,8.89593696594238)
--(axis cs:26,45.001238822937);

\path [draw=color0, semithick]
(axis cs:30,5.37409329414368)
--(axis cs:30,32.9374299049377);

\path [draw=color0, semithick]
(axis cs:34,2.56160831451416)
--(axis cs:34,37.4211483001709);

\path [draw=color0, semithick]
(axis cs:38,5.67810726165771)
--(axis cs:38,40.9954957962036);
\addplot [semithick, color0, mark=-, mark size=3, mark options={solid}, only marks]
table {%
	2 0.196285665035248
	6 0.176678530871868
	10 0.323243409395218
	14 0.660858139395714
	18 4.36999273300171
	22 16.2232275009155
	26 8.89593696594238
	30 5.37409329414368
	34 2.56160831451416
	38 5.67810726165771
};
\addplot [semithick, color0, mark=-, mark size=3, mark options={solid}, only marks]
table {%
	2 1.49478057026863
	6 1.36745424568653
	10 4.24038273096085
	14 15.5382318496704
	18 26.2786116600037
	22 54.0086536407471
	26 45.001238822937
	30 32.9374299049377
	34 37.4211483001709
	38 40.9954957962036
};
\addplot [name path=upper0,draw=none] table{%
	2 1.49478057026863
	6 1.36745424568653
	10 4.24038273096085
	14 15.5382318496704
	18 26.2786116600037
	22 54.0086536407471
};
\addplot [name path=lower0,draw=none] table{%
	2 0.196285665035248
	6 0.176678530871868
	10 0.323243409395218
	14 0.660858139395714
	18 4.36999273300171
	22 16.2232275009155
};
\addplot [fill=color0!10] fill between[of=upper0 and lower0];

\end{axis}
\begin{axis}[
height=151.86214363138544,
tick align=outside,
tick pos=left,
width=235.71811,
x grid style={white!69.0196078431373!black},
xmin=-1.8, xmax=22,
xlabel={Query travel length [m]},
ylabel={RMSE},
xtick style={color=black},
y grid style={white!69.0196078431373!black},
ymin=-2.51492022462189, ymax=56.7002523962408,
ytick style={color=black},
xshift=190.71811,
ylabel={Relative frequency},
hide x axis,
ylabel near ticks,
ymin=-0.02720285684226213, ymax=0.6133032904003803,
]

\end{axis}
\vspace{10pt}
\end{tikzpicture}
	
\begin{tikzpicture}

\definecolor{color1}{rgb}{0.3803921568627451, 0.6627450980392157, 0.3607843137254902}
\definecolor{color0}{rgb}{1.0, 0.5725490196078431, 0.1411764705882353}

\begin{axis}[
height=151.86214363138544,
tick align=outside,
tick pos=left,
width=190.71811,
x grid style={white!69.0196078431373!black},
xmin=-1.8, xmax=24,
xlabel={Query travel length [m]},
ylabel={Location error [m]},
xtick style={color=black},
y grid style={white!69.0196078431373!black},
ylabel near ticks,
ymin=-2.51492022462189, ymax=56.7002523962408,
ytick style={color=black},
legend cell align={left},
legend style={at={(0.26,0.8)},anchor=south west,fill opacity=0.8, draw opacity=1, text opacity=1, draw=white!80!black}
]
\addplot [semithick, color0]
table {%
	2 0.324848502874374
	6 0.32111394405365
	10 0.62004417181015
	14 1.06986749172211
	18 1.5120085477829
	22 2.60484218597412
	26 2.01014161109924
	30 1.31971681118011
	34 2.78060984611511
	38 2.30645227432251
};
\addlegendentry{Location error (FQ)}
\addplot+[gray,opacity=0.3, ybar interval,mark=no, fill=lightgray] 
table [x expr=\thisrow{X}, y expr=\thisrow{Y}*(56)]{%
	X Y
	0 0.99635
	4 0.00245
	8 0.00072
	12 0.00015
	16 0.00024
	20 0.00002
	24 0
};
\addlegendentry{Histogram (LQ)}

\path [draw=color0, semithick]
(axis cs:2,0.137403979897499)
--(axis cs:2,1.02253070473671);

\path [draw=color0, semithick]
(axis cs:6,0.13606370985508)
--(axis cs:6,0.996185541152954);

\path [draw=color0, semithick]
(axis cs:10,0.321723952889442)
--(axis cs:10,7.16226905584335);

\path [draw=color0, semithick]
(axis cs:14,0.654264107346535)
--(axis cs:14,14.7761322259903);

\path [draw=color0, semithick]
(axis cs:18,1.01850253343582)
--(axis cs:18,17.5050753355026);

\path [draw=color0, semithick]
(axis cs:22,2.04020315408707)
--(axis cs:22,19.7368755340576);

\addplot [semithick, color0, mark=-, mark size=3, mark options={solid}, only marks]
table {%
	2 0.137403979897499
	6 0.13606370985508
	10 0.321723952889442
	14 0.654264107346535
	18 1.01850253343582
	22 2.04020315408707
	26 1.52541226148605
	30 0.830213844776154
	34 2.19842272996902
	38 1.74839359521866
};
\addplot [semithick, color0, mark=-, mark size=3, mark options={solid}, only marks]
table {%
	2 1.02253070473671
	6 0.996185541152954
	10 7.16226905584335
	14 14.7761322259903
	18 17.5050753355026
	22 19.7368755340576
	26 21.3880217075348
	30 17.3030985593796
	34 22.802494764328
	38 20.8695645332336
};
%
\addplot [name path=upper,draw=none] table{%
	2 1.02253070473671
	6 0.996185541152954
	10 7.16226905584335
	14 14.7761322259903
	18 17.5050753355026
	22 19.7368755340576
	26 21.3880217075348
	30 17.3030985593796
	34 22.802494764328
	38 20.8695645332336
};
\addplot [name path=lower,draw=none] table{%
	2 0.137403979897499
	6 0.13606370985508
	10 0.321723952889442
	14 0.654264107346535
	18 1.01850253343582
	22 2.04020315408707
	26 1.52541226148605
	30 0.830213844776154
	34 2.19842272996902
	38 1.74839359521866
};
\addplot [fill=color0!10] fill between[of=upper and lower];
\end{axis}
\begin{axis}[
height=151.86214363138544,
tick align=outside,
tick pos=left,
width=235.71811,
x grid style={white!69.0196078431373!black},
xmin=-1.8, xmax=22,
xlabel={Query travel length [m]},
ylabel={RMSE},
xtick style={color=black},
y grid style={white!69.0196078431373!black},
ymin=-2.51492022462189, ymax=56.7002523962408,
ytick style={color=black},
xshift=190.71811,
ylabel={Relative frequency},
hide x axis,
ylabel near ticks,
ymin=-0.04490928972539089, ymax=1.0125045070757284,
]

\end{axis}
\end{tikzpicture}
	\vspace{-10pt}
	\caption{\textit{Top:} Propagation approach (no query refinement.)  \textit{Bottom:} Query refinement approach (after each layer).\\
		The median location errors of detected objects are plotted in green and orange, with $25$th and $75$th percentiles as error bars. They are displayed in terms of the distance between the \textit{first} query anchor location (FQ) and the final bounding box. The curve is flatter in the bottom plot due to anchor refinement. The effect of refinement is also visible in the histograms (gray): Here, the empirical distributions of observed travel lengths of the \textit{last} query anchor location (LQ) towards its box are depicted, with y-axis on the right.}
	\label{fig_rmse_query_movement_propagation_approach}
	\vspace{-12pt}
\end{figure}
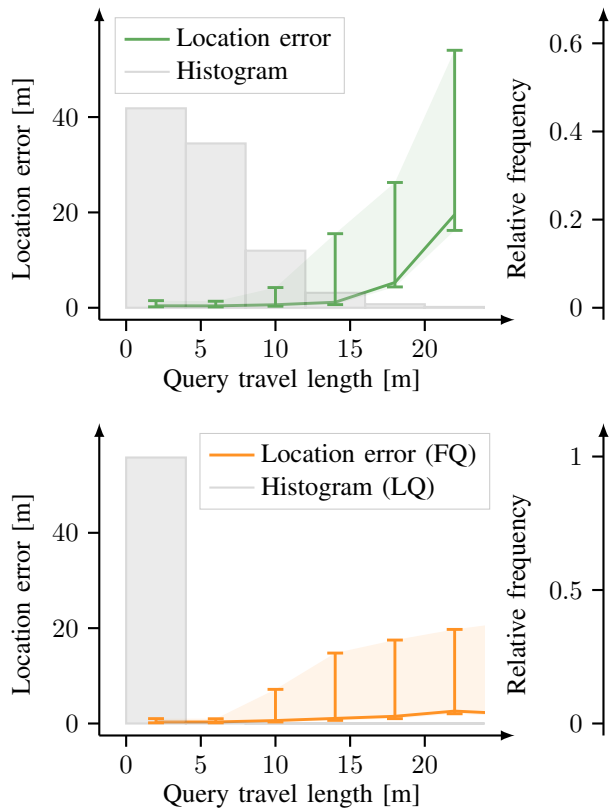
\begin{table}[h]
	\setlength\tabcolsep{0pt} 
	\caption{Query refinement variants (category car)}
	\label{tab:refinement_variants}
	\begin{tabular*}{\columnwidth}{@{\extracolsep{\fill}} ll cccc}
		\toprule
		Method& $S_\textrm{r}$  & AP$\uparrow$ & ATE (m)$\downarrow$& ASE (1-IOU) $\downarrow$& AOE (rad)$\downarrow$\\
		\midrule
		Propagation &$\{\varnothing\}$  &0.713&	0.310&	0.158&0.086 \\
		Once &$\{1\}$  &0.750&0.260&0.157&0.082 \\
		Every $2$nd&$\{1,3,5\}$   &0.756&	0.255&	0.157&	0.081 \\
		After each&$\{1,2,3,4,5\}$   &\textbf{0.759}&\textbf{0.251}&\textbf{0.157}&\textbf{0.080} \\
		\bottomrule
		
	\end{tabular*}
\vspace{-10pt}
\end{table}
\subsection{Non-Maximum Suppression}\label{nms}
We evaluate whether non-maximum suppression (NMS), i.e. the suppression of overlapping box estimates, can improve the detection results. Since the estimation of duplicate boxes is punished during training by assigning them to the 'no-object' class, and because the queries are able to communicate with one another during decoder \textit{self-attention}, it is expected that the effect of NMS should be small. The results confirm this expectation, as presented in Table \ref{tab:nms}. Different values for the minimum overlap of two boxes in order to be affected by NMS are examined ($\cap$-symbol). The observed improvement due to NMS is so subtle that we conclude that is not necessary for TransLPC.
\begin{table}[h]
	\setlength\tabcolsep{0pt} 
	\caption{Non-maximum suppression analysis (category car)}
	\label{tab:nms}
	\begin{tabular*}{\columnwidth}{@{\extracolsep{\fill}} lc cccc}
		\toprule
		&$\cap$  & AP$\uparrow$ & ATE (m)$\downarrow$& ASE (1-IOU) $\downarrow$& AOE (rad)$\downarrow$\\
		\midrule
		No NMS &-&0.759&0.251&0.157&0.080\\
		NMS  &0.1&0.759&	\textbf{0.245}&	\textbf{0.156}&	\textbf{0.077} \\
		NMS  &0.2&\textbf{0.762}&	0.247&	0.157&	0.079 \\
		
		\bottomrule
		
	\end{tabular*}
\vspace{-1pt}
\end{table}
\section{CONCLUSION}
We presented TransLPC, a novel transformer-based detection model that is applicable to large point clouds. Such point clouds are challenging due to the memory constraints in a transformer model. Our model alleviates these issues, allowing for suitably large input lengths. With the proposed query refinement technique, TransLPC improves upon existing methods in terms of estimation accuracy. Besides this, we provided insights into the transformer query behavior, both with and without refinement.

The model is modular in terms of the query refinement placement and of the backbone choice. Since it utilizes some standard components, TransLPC will directly benefit from the current surge of research interest in transformers, as well as in object detection. 

The proposed model enables the usage of transformers for further downstream tasks, which require object detection and that are supposed to operate on large point clouds, for example in autonomous driving. The vectors that are output by the model's decoder are interpretable as object-bound feature vectors and therefore useful for tasks such as tracking \cite{meinhardt_trackformer:_2021, ruppel2022trans} and prediction. Besides this, the model could be extended to a multi-modal input, for which the transformer is a natural fit \cite{hu_transformer_2021}.





\bibliographystyle{IEEEtran}

\bibliography{IEEEabrv,references}

\end{document}